\documentclass[tikz]{article}

\usepackage{PRIMEarxiv}

\usepackage[utf8]{inputenc} 
\usepackage[T1]{fontenc}    
\usepackage{hyperref}       
\usepackage{url}            
\usepackage{booktabs}       
\usepackage{amsfonts}       
\usepackage{nicefrac}       
\usepackage{microtype}      
\usepackage{lipsum}
\usepackage{fancyhdr}       
\usepackage{graphicx}       
\usepackage{csquotes}
\usepackage{mathtools} 
\usepackage{booktabs} 
\usepackage{algorithm}
\usepackage[noend]{algorithmic}
\usepackage{multirow}
\usepackage{caption}
\usepackage{subcaption}
\usepackage{amssymb}
\usepackage{graphicx}
\usepackage{xfrac}
\usepackage{xcolor}
\usepackage{ulem}
\usepackage{microtype}
\usepackage{hyperref}
\usepackage[acronym]{glossaries}

\usepackage{cleveref}

\title{
    Enhancing Decision Space Diversity in Multi-Objective Evolutionary Optimization for the Diet Problem
}

\author{
Gustavo V. Nascimento$^1$, Ivan R. Meneghini$^1$, Valéria Santos$^1$, \\ 
\textbf{Eduardo Luz$^1$ and Gladston Moreira$^1$}\\
 $^1$Computing Department, Universidade Federal de Ouro Preto, Ouro
Preto, 35402-136, Minas Gerais, Brazil.\\
    \texttt{gladston@ufop.edu.br}
}

\begin{document}
\maketitle

\begin{abstract}
Multi-objective evolutionary algorithms (MOEAs) are essential for solving complex optimization problems, such as the diet problem, where balancing conflicting objectives, like cost and nutritional content, is crucial.
However, most MOEAs focus on optimizing solutions in the objective space, often neglecting the diversity of solutions in the decision space, which is critical for providing decision-makers with a wide range of choices.
This paper introduces an approach that directly integrates a Hamming distance-based measure of uniformity into the selection mechanism of a MOEA to enhance decision space diversity.
Experiments on a multi-objective formulation of the diet problem demonstrate that our approach significantly improves decision space diversity compared to NSGA-II, while maintaining comparable objective space performance.
The proposed method offers a generalizable strategy for integrating decision space awareness into MOEAs.
\end{abstract}


\section{Introduction}
The growing emphasis on healthy lifestyles has increased the importance of optimizing food selection to meet nutritional needs while minimizing cost and maximizing variety \cite{IFIC2023,GlobalData2020,duarte2021}.
The diet problem, initially proposed by \cite{stigler1945}, captures the core challenge of finding a cost-effective food combination that satisfies nutritional requirements.
In real-world scenarios, the diet problem extends beyond cost minimization to include multiple conflicting objectives, such as maximizing nutritional diversity and specific nutrient intake, posing a significant multi-objective optimization challenge \cite{Amin19, Marrero2020, Pochmann2022}.
The high dimensionality and combinatorial nature of the problem make finding a set of Pareto-optimal solutions computationally intensive, necessitating efficient multi-objective evolutionary algorithms (MOEAs) \cite{holland1975}.

While MOEAs excel at exploring the objective space to approximate the Pareto front, they often overlook the importance of diversity in the decision space \cite{Moreira2019}.
Offering a wide range of food combinations is crucial for effective diet planning, enabling decision-makers (e.g., nutritionists or individuals) to select solutions that best suit their personal preferences and constraints.

This paper addresses this gap by introducing the Dominance-Weight Hamming Distance (DWH) heuristic, a heuristic that explicitly promotes decision space diversity, using a Hamming distance-based uniformity measure. The DWH heuristic extends the Dominance-Weighted Uniformity (DWU) approach in \cite{Moreira2019} for discrete multi-objective optimization problems. 
We evaluate our approach on a multi-objective formulation of the diet problem, comparing it against the well-established NSGA-II \cite{Deb2002}.
Our results demonstrate a significant improvement in decision space diversity, as measured by the minimum and average Hamming distances, with comparable Hypervolume in the objective space.
This suggests that our method effectively explores a wider range of dietary options without compromising the quality of the solutions in terms of cost, protein content, and variety.
Our key contributions are:
\begin{itemize}
    \item Integration of a Hamming distance-based uniformity measure into the selection process of a MOEA to enhance decision space diversity;
    \item A demonstration of the effectiveness of this approach in the context of the multi-objective diet problem, a challenging real-world optimization problem;
    \item A quantitative comparison with NSGA-II, highlighting the trade-off between decision space diversity and objective space performance.
\end{itemize}
Future work will explore the application of this approach to other multi-objective optimization problems and investigate alternative measures of decision space diversity.

\section{Related Work}

The multi-objective diet problem has been addressed using various MOEAs and modeling techniques. In \cite{kaldirim2006}, the authors use NSGA-II to generate personalized diets based on user characteristics, focusing on objective space optimization. A multi-objective model that minimizes cost, unhealthy fats, and sugar, while maximizing fiber intake, is proposed in \cite{Amin19}. Still, their approach uses scalarization methods, which are known to sometimes struggle with Pareto front diversity.

The work in \cite{Marrero2020} presents an approach to the menu planning problem, specifically for school canteens. The approach aims to minimize the cost of the menus and the repetition of courses and food groups. To solve this, the authors proposed a multi-objective memetic algorithm based on the MOEA/D framework.

In \cite{Ramos-Perez2021}, the authors compared different MOEAs for school lunch planning, showing the effectiveness of NSGA-II and SPEA2 under different conditions. However, their analysis centers on objective space metrics, such as Hypervolume. A bi-level recommender system for food diets, utilizing NSGA-II at the higher level to determine the optimal combination of foods, is presented in \cite{Pochmann2022}. The authors in \cite{Türkmenoğlu2022} incorporated fuzzy inference to represent user preferences in the objective function, showing the importance of personalized models.

The authors in \cite{Moreira2019} introduced a uniformity measure weighted by dominance for decision space diversity.  However, its formulation aims to solve continuous multi-objective optimization problems.

This paper introduces the DWH heuristic, a heuristic based on the Hamming distance, which is more suitable for the discrete nature of the diet problem's decision variables (i.e., food choices).


\section{Background}

\subsection{Problem Setting}
The diet problem, in its basic form, aims to minimize the cost of a diet while satisfying nutritional requirements \cite{stigler1945}.
Let $A = \{1, 2, ..., n\}$ be the set of $n$ available food items, and $N = \{1, 2, ..., m\}$ be the set of $m$ essential nutrients.
The classical formulation, using linear programming, is:
\begin{equation}
    \text{minimize} \quad f_1(x) = \sum_{i=1}^{n} c_i x_i
    \label{eq:cost}
\end{equation}
subject to:
\begin{equation}
    \sum_{i=1}^{n} a_{ij} x_i \geq R_j, \quad \forall j \in N
    \label{eq:nutrition}
\end{equation}
\begin{equation}
    x_i \geq 0, \quad \forall i \in A
    \label{eq:nonneg}
\end{equation}
where $c_i$ is the cost per serving of food item $i$, $x_i$ is the number of servings of food item $i$, $a_{ij}$ is the amount of nutrient $j$ in food item $i$, and $R_j$ is the minimum required amount of nutrient $j$.

In this work, we extend this classical formulation to a multi-objective optimization problem over a 7-day period, considering the following objectives:

\begin{enumerate}
    \item  \textbf{Minimize Cost:}  $f_1$ as defined in Equation \ref{eq:cost}, extended over 7 days:
    \begin{equation}
        f_1(x) = \sum_{d=1}^{7} \sum_{i=1}^{n} c_i x_{id}
        \label{eq:cost_multi_day}
    \end{equation}
    where $x_{id}$ is the number of servings of food item $i$ on day $d$.

    \item  \textbf{Minimize Repetitiveness:}  Following \cite{Ramos-Perez2021}, we minimize the repetitiveness of food categories across days.
    Let $G = \{1, 2, ..., p\}$ be the set of $p$ food categories.
    The repetitiveness penalty for day $d$ is:
    \begin{equation}
        v_d = \sum_{k=1}^{d-1} \left( \sum_{g=1}^{p} y_{gk} p_g + z_k p_{p+k} \right)
        \label{eq:repetitiveness_day}
    \end{equation}
    where $y_{gk} = 1$ if category $g$ is repeated on day $d-k$ and 0 otherwise, $z_k = 1$ if any category is repeated $k$ days before, and $p_g$ and $p_{p+k}$ are the repetition penalties (see Table \ref{tab:penalties}).
    The overall repetitiveness objective is:
    \begin{equation}
        f_2(x) = \sum_{d=1}^{7} v_d
        \label{eq:repetitiveness_overall}
    \end{equation}

    \item  \textbf{Maximize Protein:} We maximize the total protein content of the diet:
    \begin{equation}
        f_3(x) = \sum_{d=1}^{7} \sum_{i=1}^{n} u_i x_{id}
        \label{eq:protein}
    \end{equation}
    where $u_i$ is the protein content per serving of food item $i$.
\end{enumerate}

The multi-objective problem is then:
\begin{equation}
    \text{minimize} \quad F(x) = (f_1(x), f_2(x), -f_3(x))
    \label{eq:multi_objective}
\end{equation}
subject to:
\begin{equation}
    \sum_{i=1}^{n} a_{ij} x_{id} \geq R_j, \quad \forall j \in N, \forall d \in \{1, 2, ..., 7\}
    \label{eq:nutrition_multi_day}
\end{equation}
\begin{equation}
    x_{id} \geq 0, \quad \forall i \in A, \forall d \in \{1, 2, ..., 7\}
    \label{eq:nonneg_multi_day}
\end{equation}

This formulation presents a challenging multi-objective optimization problem with a large, complex search space. Our approach focuses on effectively exploring this space while maintaining diversity in the set of solutions.
\begin{table}[!ht]
\caption{Repetition penalties considered to minimize the repetitiveness of food categories across days.}
\label{tab:penalties}
\begin{center}
\begin{tabular}{clc}
\toprule
\textbf{Penalty} & \textbf{Description} & \textbf{Value} \\
\midrule
$p_1$ & Repeat ``Other'' category & 0.1 \\
$p_2$ & Repeat ``Meats'' category & 3 \\
$p_3$ & Repeat ``Cereals'' category & 0.3 \\
$p_4$ & Repeat ``Fruits'' category & 0.1 \\
$p_5$ & Repeat ``Dairy'' category & 0.3 \\
$p_6$ & Repeat ``Legumes'' category & 0.3 \\
$p_7$ & Repeat ``Seafood'' category & 0.5 \\
$p_8$ & Repeat ``Vegetables'' category & 0.1 \\
$p_9$ & Repeat category 1 day before & 3 \\
$p_{10}$ & Repeat category 2 days before & 2.5 \\
$p_{11}$ & Repeat category 3 days before & 1.8 \\
$p_{12}$ & Repeat category 4 days before & 1 \\
$p_{13}$ & Repeat category 5 days before & 0.2 \\
$p_{14}$ & Repeat category 6 days before & 0.1 \\
\bottomrule
\end{tabular}
\end{center}
\end{table}

\subsection{Dataset}

We used the TACO dataset - (Brazilian Food Composition Table)\cite{taco2011}, which provides nutritional information for 597 food items across 15 categories (see Table \ref{tab:categories}).
Since the TACO dataset lacks cost information, we assigned random costs between \$1.00 and \$10.00 to each food item.
The nutritional requirements ($R_j$ in Equation \eqref{eq:nutrition_multi_day}) were based on the Brazilian Recommended Daily Intake values provided by ANVISA - (National Health Surveillance Agency in Brazil)\cite{anvisa_in75_2020}.
\begin{table}[!ht]
\caption{TACO dataset \cite{taco2011} composed of 15 Food Categories with 597 food items.}
\label{tab:categories}
\begin{center}
\begin{tabular}{lc}
\toprule
\textbf{Category} & \textbf{Number of Items} \\
\midrule
Cereals and derivatives& 63 \\
Vegetables, greens, and derivatives& 99 \\
Fruits and derivatives& 96 \\
Fats and oils & 14 \\
fish and Seafood & 50 \\
Meats and meat products& 123 \\
Milk and dairy products& 24 \\
Alcoholic and non-alcoholic beverages & 14 \\
Eggs and derivatives& 7 \\
Sugary products& 20 \\
Miscellaneous & 9 \\
Other processed foods & 5 \\
Prepared food& 32 \\
Legumes and derivatives& 30 \\
Nuts and seeds & 11 \\
\bottomrule
\end{tabular}
\end{center}
\end{table}

\section{Method}

Our approach integrates a Hamming distance-based uniformity measure into the selection process of a multi-objective evolutionary algorithm.
The aim is to generate a set of Pareto-optimal solutions that are well-distributed in the objective space and diverse in the decision space.

\subsection{Uniformity Measure based on Hamming Distance}

We define a uniformity measure, $w_{dH}$, that quantifies the dissimilarity between two diets based on their Hamming distance.
Given two diets $\mathcal{D}$ and $\mathcal{D}'$, represented by binary vectors $x$ and $x'$ of length $n$ (the number of food items), $w_{dH}$ is calculated as:
\begin{equation}
    w_{dH}(x, x') = \frac{d_H(x, x')}{|r(x) - r(x')| + 1}
    \label{eq:uniformity_hamming}
\end{equation}
where $d_H(x, x')$ is the Hamming distance between the binary vectors representing the diets, defined by equation \eqref{eq:hamming}, and $r(x)$ is the SPEA-2 raw fitness of solution $x$, as described in \cite{Zitzler2001}.
The denominator normalizes the Hamming distance by the difference in fitness values, giving more weight to solutions that are dissimilar in the decision space and have similar fitness.
This measure is then incorporated into a MinMax heuristic, shown in Algorithm \ref{alg:DWH}, to guide the selection process in the evolutionary algorithm.

\subsection{Multi-Objective Evolutionary Algorithm with Hamming Diversity (MOEA-HD)}

Our MOEA-HD algorithm follows a standard evolutionary algorithm framework with modifications to the selection process to incorporate the $w_{dH}$ measure.
A solution is represented as a matrix of size $n \times 7$, where $n$ is the number of food items, and each element $(i, d)$ represents the number of servings of food item $i$ on day $d$.
The algorithm proceeds as follows:

\begin{enumerate}
    \item  \textbf{Initialization:} Generate an initial population $P_0$ of $k$ individuals (diets) randomly. Apply a repair operator to ensure all individuals satisfy the nutritional constraints (Equation \ref{eq:nutrition_multi_day}).
    \item  \textbf{Iteration:} For $t = 0$ to $max\_gen$:
    \begin{enumerate}
        \item  \textbf{Fitness Assignment:} Calculate the fitness of each individual in $P_t$ using Non-dominated Sorting as in NSGA-II \cite{Deb2002}.
        \item  \textbf{Selection:} Perform $k$ binary tournaments. The winner of each tournament is determined by:
        \begin{itemize}
            \item If one solution is feasible and the other is not, the feasible solution wins.
            \item If both are infeasible, the solution with the lower constraint violation penalty (Equation \ref{eq:penalty}) wins.
            \item If both are feasible, the solution that maximizes $w_{dH}$ with respect to other selected solutions wins (using Algorithm \ref{alg:DWH}).
        \end{itemize}
        \item  \textbf{Crossover:} Apply multi-point crossover (20 points) on selected pairs to generate offspring.
        \item  \textbf{Mutation:} Apply bit-flip mutation to the offspring.
        \item  \textbf{Population Update:} Combine $P_t$ and the offspring, and use Algorithm \ref{alg:DWH} to select the $k$ individuals for $P_{t+1}$.
    \end{enumerate}
    \item  \textbf{Final Repair:} Apply the repair operator to the individuals in the final population $P_{max\_gen}$.
    \item  \textbf{Output:} Return the non-dominated solutions from $P_{max\_gen}$.
\end{enumerate}

\begin{algorithm}
\caption{DWH Heuristic for Diversity Selection}
\label{alg:DWH}
\begin{algorithmic}[1]
\REQUIRE Population $P$, scalar $k$
\ENSURE Set of diverse solutions $R$
\STATE $N_P \leftarrow$ Non-dominated solutions of $P$
\STATE $R \leftarrow \arg\max_{x, x' \in N_P} w_{dH}(x, x')$
\WHILE{$|R| < k$}
\STATE $x' \leftarrow \arg\max_{x \in P \setminus R} \min_{r \in R} w_{dH}(x, r)$
\STATE $R \leftarrow R \cup \{x'\}$
\ENDWHILE
\RETURN $R$
\end{algorithmic}
\end{algorithm}

\subsection{Constraint Handling}

Solutions that violate the nutritional constraints (Equation \ref{eq:nutrition_multi_day}) are penalized.
The penalty is calculated as the sum of the deficits for each nutrient and day:
\begin{equation}
    \text{penalty} = \sum_{d=1}^{7} \sum_{j=1}^{m} \max(0, R_j - a_{ij} x_{id})
    \label{eq:penalty}
\end{equation}
This penalty is used in the selection process to favor feasible solutions and to differentiate between infeasible solutions.

\section{Experimental Setup}

To evaluate the effectiveness of our MOEA-HD, we compared its performance with that of NSGA-II \cite{Deb2002}, a widely used and effective multi-objective evolutionary algorithm.
Both algorithms were implemented in Python, and experiments were conducted on a machine equipped with a Ryzen 5 2600 processor and 16GB of RAM.

\subsection{Hamming Distance}
The Hamming distance \cite{hamming1950} is a metric used to measure the difference between two binary strings.
Given two strings of equal length, the Hamming distance is the number of positions at which the corresponding symbols differ.
Formally, given two binary strings $a = a_1a_2 \ldots a_n$ and $b = b_1b_2 \ldots b_n$, with $a_i, b_j \in \{0, 1\}, ~ i, j = 1, 2, \ldots n$, the Hamming distance $d_H$ between $a$ and $b$ is:
\begin{equation}
    d_H(a, b) = \sum_{i=1}^{n} |a_i - b_i|
    \label{eq:hamming}
\end{equation}

In this work we will represent the string $x=x_1x_2, \ldots x_n$ as a binary vector $x = (x_1,x_2, \ldots, x_n), ~ x_i \in \{0,1\}, ~ i = 1, 2, \ldots n$. Furthermore, we represent each diet as a binary vector where each element indicates the presence or absence of a particular food item. The Hamming distance then measures how different two diets are in terms of their food composition.

\subsection{Performance Metrics}
To evaluate the performance of our algorithm, we use two sets of metrics:

\subsubsection{Objective Space Metric}
\begin{itemize}
    \item \textbf{Hypervolume:}  This metric measures the region's volume in the objective space dominated by the non-dominated solutions and bounded by a reference point \cite{Zitzler}. A higher hypervolume generally indicates a better approximation of the Pareto front.
\end{itemize}

\subsubsection{Decision Space Metrics} \label{sec:metric}
\begin{itemize}
    \item $\mathbf{d_{Hmin}}$: The minimum Hamming distance between any two solutions in the non-dominated set. This metric reflects the closeness of the most similar solutions; a higher value indicates better diversity.
\begin{equation} \label{eq:dhmin}
d_{Hmin} = min(d_H(x,x'))
\end{equation}
    \item $\mathbf{d_{Hmed}}$: The average Hamming distance between all pairs of solutions in the non-dominated set. This provides an overall measure of diversity in the decision space; a higher value is desirable.
    \begin{equation}\label{eq:dhmed}
    d_{Hmed} = \frac{\sum_{x' \in \mathcal{D}}\sum_{x \in \mathcal{D}}d_H(x,x')}{|\mathcal{D}|^2}
\end{equation}
\end{itemize}

\subsection{Parameters}

The parameters for both algorithms were:

\begin{itemize}
    \item Population size: 30
    \item Initialization: Food quantities were randomly initialized with a weighted distribution (94\% chance of 0, 4\% of 1, 1\% of 2 servings).
    \item Maximum generations: 30, 100, 300
    \item Crossover: 20-point multi-point crossover
    \item Mutation: Bit-flip with 5\% probability for 1-to-0 and 0.16\% for 0-to-1 flips.
\end{itemize}

For each algorithm, we run it 30 times for each generation setting, and the average results were recorded.
Hypervolume was calculated using a reference point based on the worst-case objective values, and the objective values were normalized prior to hypervolume calculation.

\section{Results and Discussion}

\subsection{Performance metrics}
Table \ref{tab:comparison} presents the comparative results of MOEA-HD and NSGA-II. The key observation is that MOEA-HD consistently outperforms NSGA-II in decision space diversity metrics \ref{sec:metric}.
%
This confirms our hypothesis that incorporating Hamming distance into the selection process effectively promotes the generation of more diverse dietary solutions.
\begin{table}[!ht]
\caption{Hypervolume, minimum and average Hamming distances of the solutions obtained by the MOEA-HD and NSGA-II algorithms.}
\label{tab:comparison}
\centering
\resizebox{.7\linewidth}{!}{
\begin{tabular}{lcccccc}
\toprule
\multirow{2}{*}{\textbf{Generations}} & \multicolumn{3}{c}{\textbf{MOEA-HD}} & \multicolumn{3}{c}{\textbf{NSGA-II}} \\
\cmidrule(lr){2-4} \cmidrule(lr){5-7}
& \textbf{Hv} & $\mathbf{d_{Hmin}}$ & $\mathbf{d_{Hmed}}$ & \textbf{Hv} & $\mathbf{d_{Hmin}}$ & $\mathbf{d_{Hmed}}$ \\
\midrule
30 & 0.379 & 95.03 & 184.77 & 0.446 & 43.30 & 157.90 \\
100 & 0.387 & 87.90 & 159.91 & 0.412 & 32.86 & 121.30 \\
300 & 0.389 & 90.46 & 154.39 & 0.381 & 26.56 & 108.82 \\
\bottomrule
\end{tabular}}\vspace{-.61em}
\end{table}

Specifically, at 30 generations, the minimum Hamming distance $d_{Hmin}$ for MOEA-HD is 95.03, compared to 43.30 for NSGA-II.
This indicates that even the most similar solutions generated by MOEA-HD are significantly more different than those generated by NSGA-II.
The average Hamming distance $d_{Hmed}$ also exhibits a similar trend, with MOEA-HD achieving higher values, which confirms greater overall diversity.
Importantly, this enhanced diversity in the decision space is achieved without a significant compromise in objective space performance, as the hypervolume values are comparable between the two algorithms.

\subsection{Objective space}

\begin{figure}[!ht]
    \centering
    \begin{subfigure}{0.48\textwidth}
    \includegraphics[width=\textwidth]{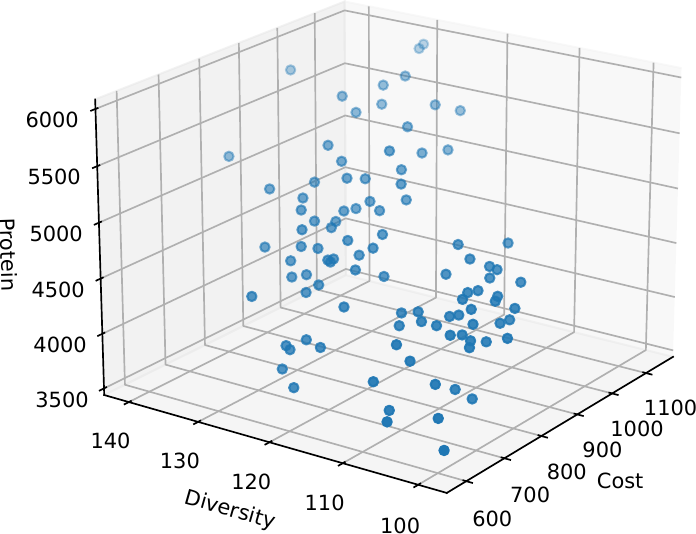}
        \caption{MOEA-HD algorithm.}
        \label{fig:dwh_pareto}
    \end{subfigure}
    \begin{subfigure}{0.48\textwidth}
        \includegraphics[width=\textwidth]{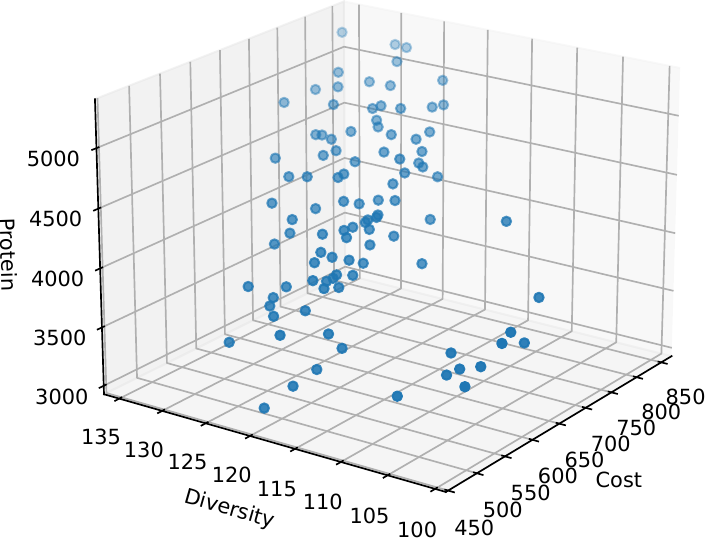}
        \caption{NSGA-II algorithm.}
        \label{fig:nsga_ii_pareto}
    \end{subfigure}    
    \caption{Pareto front approximations for Maximum generations $=300$ for: (a) MOEA-HD algorithm; (b) NSGA-II algorithm.}
    \label{fig:pareto_fronts}
\end{figure}

Figure \ref{fig:pareto_fronts} visually compares the Pareto front approximations obtained by both algorithms after 100 generations.
While both algorithms cover a similar region in the objective space, the solutions generated by MOEA-HD appear to be more evenly distributed, as reflected in the quantitative diversity measures.

\subsection{Decision space}

\begin{figure}[!ht]
    \begin{center}
    \begin{subfigure}{0.8\textwidth}
        \includegraphics[width=\textwidth]{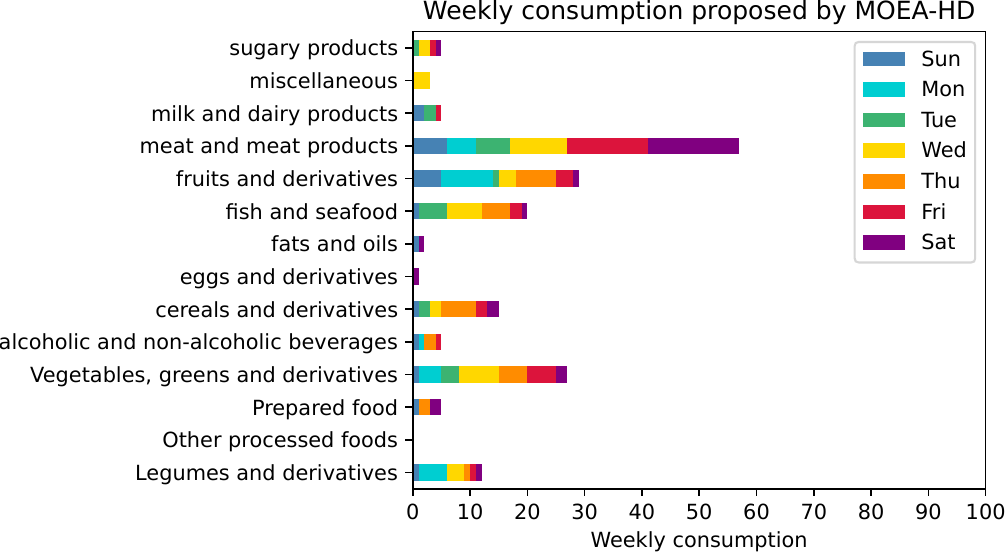}
        \caption{Items demanded by the MOEA-HD algorithm.}
        \label{fig:consumo_dwu}
    \end{subfigure}
    \end{center}
    \vspace{1em}
    \begin{center}
    \begin{subfigure}{0.8\textwidth}
        \includegraphics[width=\textwidth]{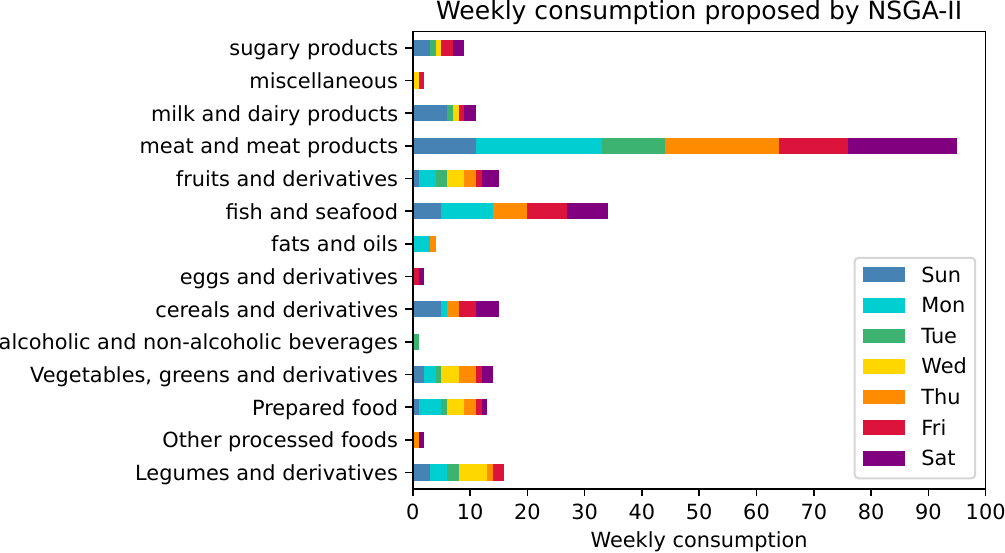}
        \caption{Items demanded by the NSGA-II algorithm.}
        \label{fig:consumo_nsga}
    \end{subfigure}
    \end{center}
    \caption{Weekly consumption of the items proposed by the algorithms.}
    \label{fig:consumo}
\end{figure}

\begin{figure}[!ht]
    \begin{center}
    \begin{subfigure}{0.8\textwidth}
        \includegraphics[width=\textwidth]{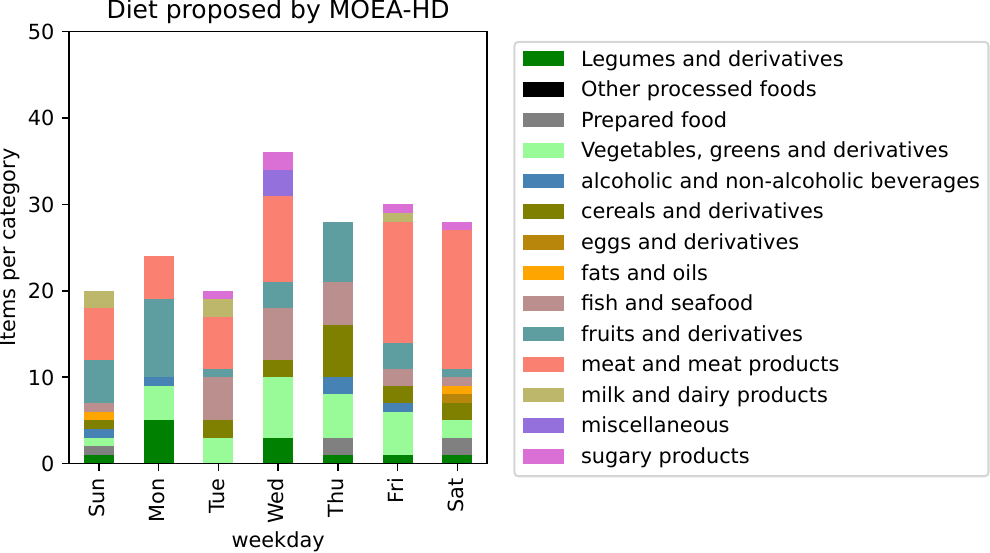}
        \caption{Items on the days of the week obtained by the MOEA-HD algorithm.}
        \label{fig:dieta_dwu}
    \end{subfigure}
    \end{center}
     \vspace{1em}
     \begin{center}
    \begin{subfigure}{0.8\textwidth}
        \includegraphics[width=\textwidth]{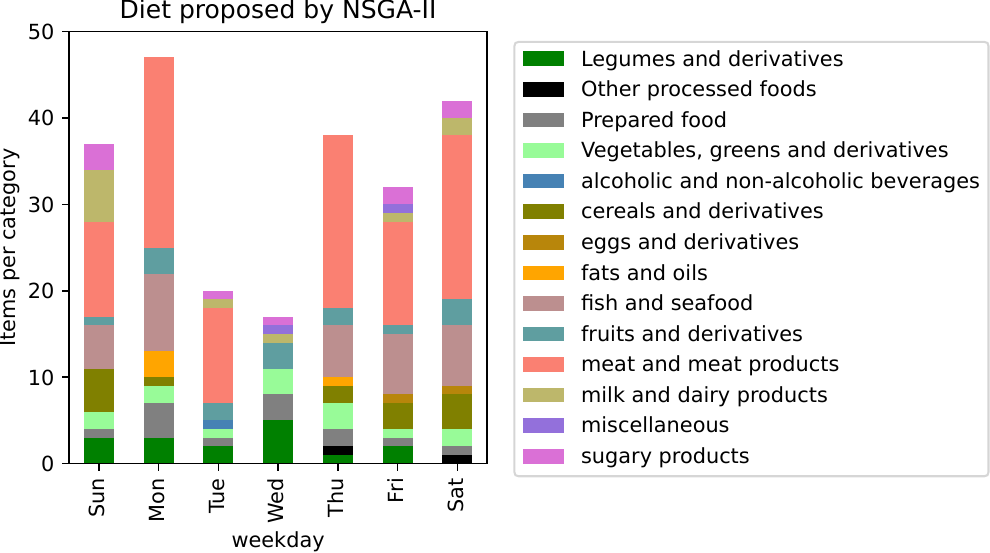}
        \caption{Items on the days of the week obtained by the NSGA-II algorithm.}
        \label{fig:dieta_nsgaii}
    \end{subfigure}
    \end{center}
    \caption{Diets proposed for each day of the week.}
    \label{fig:dieta}
    
\end{figure}

Figures \ref{fig:consumo} and \ref{fig:dieta} illustrate the diets proposed by the algorithms. In each figure, the points in the decision space of one execution of each algorithm are grouped according to the categories presented in Table \ref{tab:categories}. Figure \ref{fig:consumo} illustrates the weekly consumption of the items, while Figure \ref{fig:dieta} presents the diets proposed for each day of the week.

The comparison of Figures \ref{fig:consumo_dwu} and \ref{fig:consumo_nsga} shows that the MOEA-HD algorithm reduced the consumption of the item \textit{meat and meat products} and increased the consumption of \textit{fish and seafood}. An increase in demand for the item \textit{fruits and derivatives} and the item \textit{vegetables, greens and derivatives} is also observed in the response presented by the MOEA-HD algorithm. There was also a slight reduction in \textit{sugary products}, \textit{milk and dairy products}, and \textit{prepared foods}. 
The demand for \textit{processed food} was eliminated. No significant variation in the quantity of the items \textit{fats and oils}, \textit{eggs and derivatives}, and \textit{vegetables and derivatives} was observed. The observed variations can facilitate the purchasing process by presenting a more diversified shopping list, thereby reducing the impact of potential supply restrictions on specific items. 
In addition, a more diversified demand allows the decision maker to purchase items that are more abundant at certain times of the year, resulting in cost savings.

The use of items on the days of the week, shown in Figures \ref{fig:dieta_dwu} and \ref{fig:dieta_nsgaii}, also reveals important details. Initially, a more even distribution of item consumption is observed during a week, as indicated by the greater regularity in the height of the bars in each graph. There is an offer of animal protein (meat or fish) on all days of the week in the diets proposed by the MOEA-HD algorithm, while this item is absent on Wednesdays in all diets proposed by the NSGA-II algorithm. Another important aspect is the increase in dietary diversity proposed by the MOEA-HD algorithm, particularly on weekends, such as Saturdays and Sundays.

In general, figures \ref{fig:consumo} and \ref{fig:dieta} show that the diversification of solutions in the decision space increased the quality of the proposed solutions, both from a financial and nutritional point of view.

\subsection{Statistical validation}

We performed a Monte Carlo simulation to further validate our results' statistical significance.
For each performance metric (Hypervolume, $d_{Hmin}$, and $d_{Hmed}$) and each generation setting, we generated 5000 random permutations of the combined results from both algorithms. We calculated the difference in means for each permutation.
This allowed us to construct a distribution of mean differences under the null hypothesis that no significant difference exists between the algorithms.
We then compared the observed mean difference to this distribution to assess its statistical significance.
%
%
\begin{figure}[!ht]
    \centering
    \includegraphics[width=.8\textwidth]{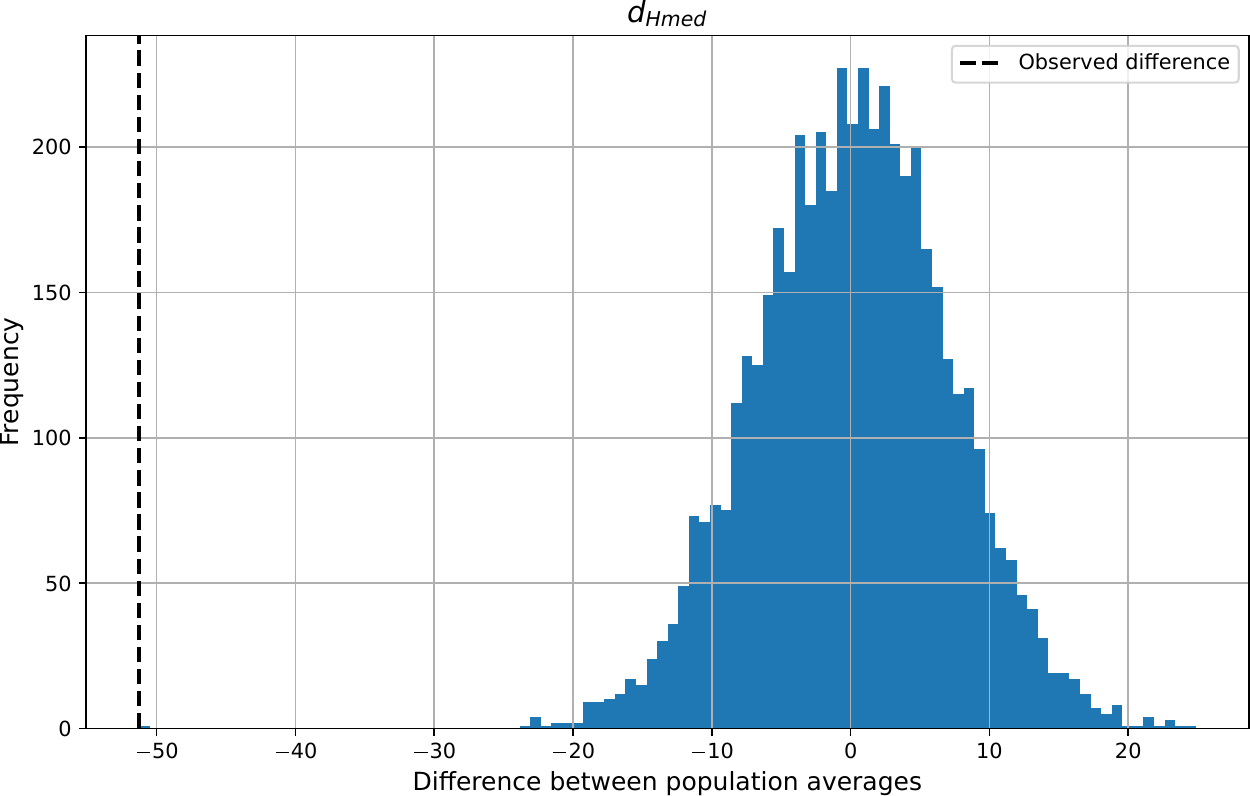}
    \caption{Monte Carlo simulation for $d_{Hmed}$ after 300 generations. The observed differences are highly statistically significant.}
    \label{fig:mc_dmed}
\end{figure}

\begin{figure}[!ht]
\centering
    \includegraphics[width=.8\textwidth]{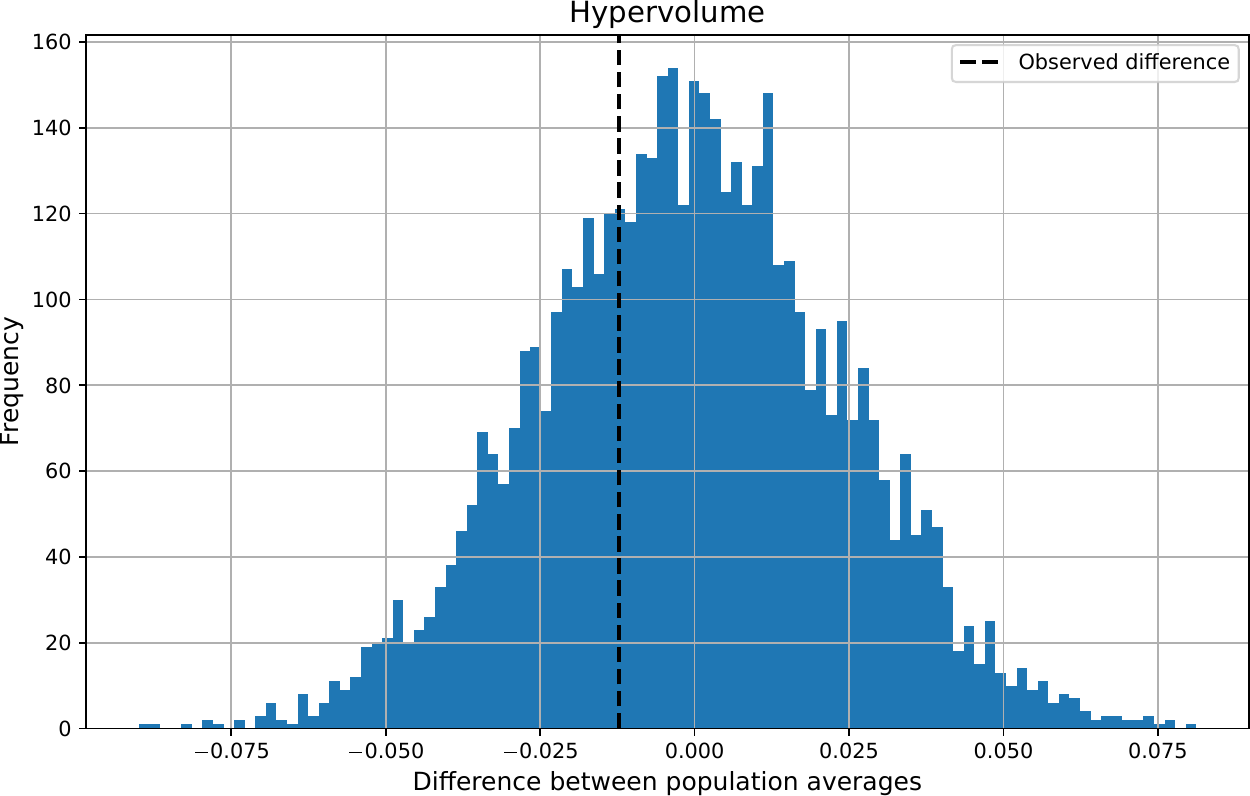}        
    \caption{Monte Carlo simulation for Hypervolume after 300 generations. The observed mean differences are not statistically significant.}
    \label{fig:mc_hv}
\end{figure}

\begin{figure}[!ht]
    \centering 
    \includegraphics[width=.8\textwidth]{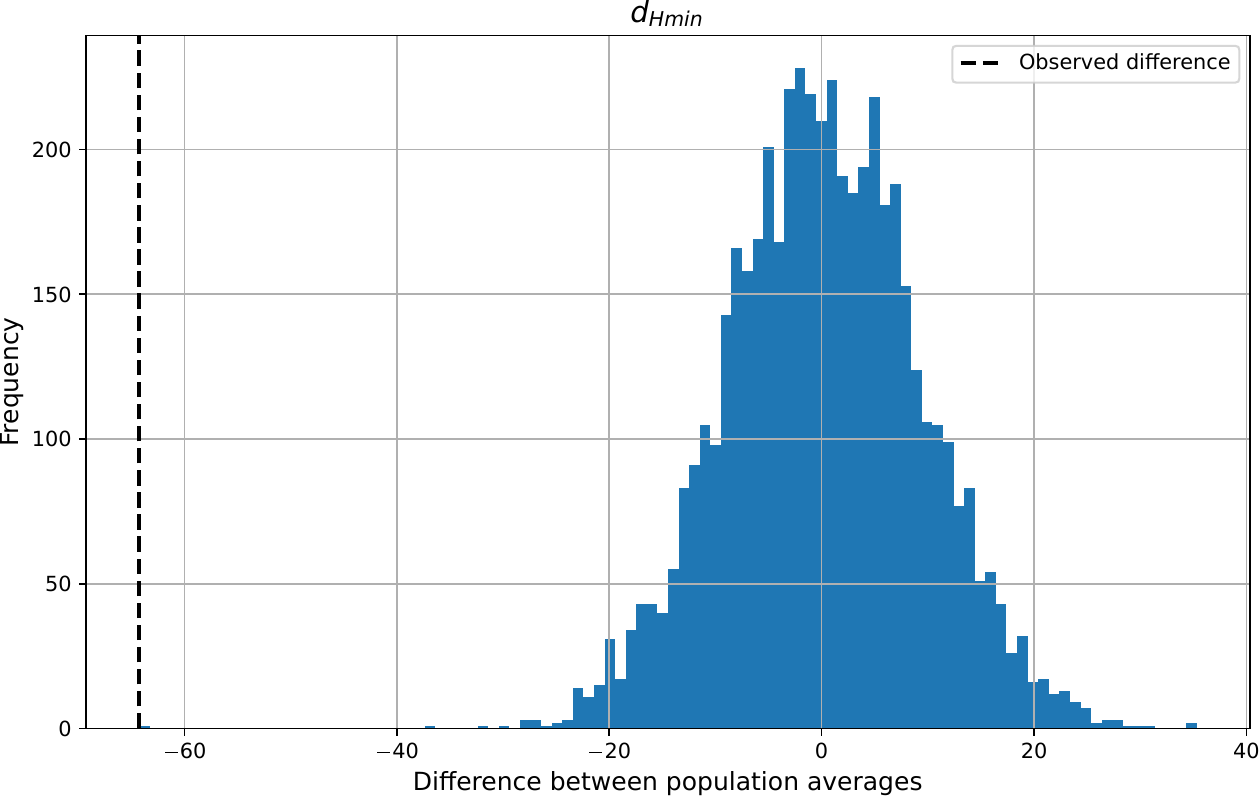} 
    \caption{Monte Carlo simulation for $d_{Hmin}$ after 300 generations. The observed differences are highly statistically significant.}
    \label{fig:mc_dmin}
\end{figure}

The results of the Monte Carlo simulations are shown in Figures \ref{fig:mc_dmed}, \ref{fig:mc_hv}, and \ref{fig:mc_dmin}.
For Hypervolume (Figure \ref{fig:mc_hv}), the observed mean differences are not statistically significant, indicating that both algorithms achieve comparable objective space performance.
However, for $d_{Hmin}$ (Figure \ref{fig:mc_dmin}) and $d_{Hmed}$ (Figure \ref{fig:mc_dmed}), the observed differences are highly significant ($p < 0.01$) across all generation settings, strongly supporting the superiority of MOEA-HD in terms of decision space diversity.

These results highlight a crucial trade-off in multi-objective optimization.
While NSGA-II excels at converging to a well-performing Pareto front in the objective space, it may sacrifice diversity in the solutions.
In contrast, MOEA-HD prioritizes exploring a wider range of solutions, providing decision-makers with more options, which can be particularly valuable in real-world applications, such as personalized diet planning.

\subsection{Limitations}

The limitations of our study include the use of randomly generated costs for food items, which may not accurately reflect actual price variations in the real world.
Future work should incorporate more realistic cost data and explore the impact of different diversity measures and selection strategies.

\section{Conclusion}

This paper presented MOEA-HD, a novel multi-objective evolutionary algorithm that enhances decision space diversity by integrating a Hamming distance-based uniformity measure into its selection mechanism.
Experiments on a multi-objective formulation of the diet problem demonstrated that MOEA-HD significantly outperforms NSGA-II in terms of decision space diversity, while maintaining comparable objective space performance. Additionally, demonstrate that the diversification of solutions in the decision space enhances the quality of the proposed solutions, both from a financial and nutritional perspective.
This suggests that MOEA-HD is a valuable tool for applications where providing diverse solutions is crucial for decision-making.
Future work will extend this approach to other multi-objective optimization problems and explore alternative diversity measures.

\section*{Acknowledgments}

This work was supported by the Conselho Nacional de Desenvolvimento Científico e Tecnológico (CNPq, grants 307151/2022-0, 308400/2022-4, 152613/2024-2), Fundação de Amparo à Pesquisa do Estado de Minas Gerais (FAPEMIG, grant APQ-01647-22). We also thank the Universidade Federal de Ouro Preto (UFOP) for their invaluable support.





\bibliographystyle{main}  
\bibliography{main}

\end{document}